\documentclass[10pt,twocolumn,letterpaper]{article}

\usepackage[pagenumbers]{cvpr} 

\usepackage{multirow}
\usepackage{amsmath, amssymb}
\usepackage[dvipsnames,table]{xcolor}
\usepackage{graphicx}
\usepackage{booktabs}
\usepackage{makecell}
\usepackage{array}
\usepackage{wrapfig}
\usepackage{bbm}
\usepackage{float}
\usepackage{perpage}
\usepackage{enumitem}
\usepackage[ruled,vlined]{algorithm2e}
\usepackage{algpseudocode}
\usepackage{extarrows}
\usepackage{marvosym}
\usepackage{pifont}
\usepackage{gensymb}
\usepackage{adjustbox}
\usepackage{ragged2e}
\usepackage{siunitx}
\usepackage{ifsym}
\usepackage{caption}

\definecolor{cvprblue}{rgb}{0.21,0.49,0.74}
\usepackage[pagebackref,breaklinks,colorlinks,citecolor=cvprblue]{hyperref}
\usepackage{microtype}

\title{Region-Aware Deformable Convolutions}

\author{
	Abolfazl Saheban Maleki \quad
	Maryam Imani \\[0.5em]
	\small
	\texttt{abolfazlmaleki1941@gmail.com} \quad
	\texttt{maryam.imani@modares.ac.ir} \\[0.8em]
	Tarbiat Modares University
}

\begin{document}
	\maketitle
	
	\begin{abstract}
We introduce Region-Aware Deformable Convolution (RAD-Conv), a new convolutional operator that enhances neural networks' ability to adapt to complex image structures. Unlike traditional deformable convolutions, which are limited to fixed quadrilateral sampling areas, RAD-Conv uses four boundary offsets per kernel element to create flexible, rectangular regions that dynamically adjust their size and shape to match image content. This approach allows precise control over the receptive field's width and height, enabling the capture of both local details and long-range dependencies, even with small 1x1 kernels. By decoupling the receptive field's shape from the kernel's structure, RAD-Conv combines the adaptability of attention mechanisms with the efficiency of standard convolutions. This innovative design offers a practical solution for building more expressive and efficient vision models, bridging the gap between rigid convolutional architectures and computationally costly attention-based methods.

\end{abstract}
	\vspace{-0.2cm}
\section{Introduction}
\label{sec:intro}

In recent years, computer vision has witnessed a rapid shift toward Transformer architectures~\cite{vaswani2017attention,dosovitskiy2020image,liu2021swin,liu2022swin,wang2021pyramid,wang2022pvt}, with large-scale foundation models leveraging attention to achieve state-of-the-art performance in classification~\cite{dosovitskiy2020image,dehghani2023scaling,dong2022cswin,yu2022coca}, detection~\cite{carion2020end,zhu2020deformable,zhang2022dino,zhao2024detrs,xia2024vit,zhang2025open,fu2025llmdet}, and segmentation~\cite{strudel2021segmenter,kirillov2023segment,xie2021segformer,cheng2022masked,bao2021beit,wang2025dual,kerssies2025your,braso2025native}. Despite their success, attention mechanisms are computationally expensive and often struggle to capture fine-grained local structures efficiently~\cite{saha2025vision}. In parallel, convolutional architectures have been revisited as competitive alternatives, preserving inductive biases such as translation equivariance and efficiency~\cite{liu2022convnet,dcnv3,ding2022replknet}. Deformable convolutions~\cite{dcnv1,dcnv2,dcnv3,dcnv4} extend this paradigm by adaptively sampling spatial locations with continuous offsets through bilinear interpolation, enabling a single layer to model long-range dependencies within fixed quadrilateral receptive fields. However, these fields remain constrained to quadrilateral geometry defined by exactly four neighboring pixels per sampling point, limiting their ability to dynamically adjust region shape and extent according to content. In contrast, RAD-Conv enables input-dependent adjustment of both region geometry (extent and aspect ratio) through boundary offsets, allowing receptive fields to dynamically conform to object structures while maintaining computational efficiency.

\begin{figure}[t]
	\centering
	\includegraphics[width=0.85\columnwidth]{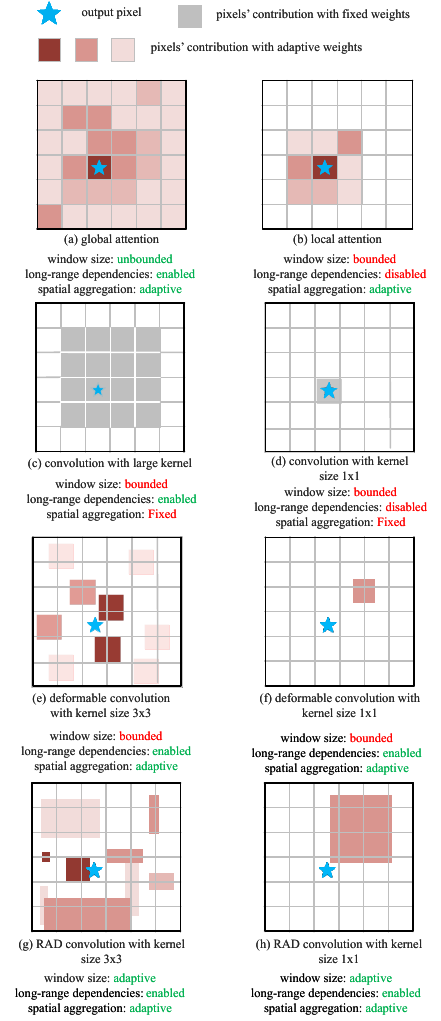}
	\caption{Comparison of vision operators by \textbf{window size} (bounded/adaptively-bounded), \textbf{long-range dependency} (enabled/disabled), and \textbf{spatial aggregation} (adaptive/fixed). Integer pixels (gray grids): pink denotes adaptive, gray fixed-weight aggregation. Global attention covers all positions at quadratic cost; local attention is window-limited. Large-kernel convolutions use fixed windows; $1\times1$ convolutions lack spatial modeling. Deformable convolutions adapt sampling via bilinear interpolation (4 pixels per point) but remain bounded by $4K$, where $K$ is kernel size. RAD-Conv enables adaptive window sizing up to image dimensions via continuous rectangular integration, achieving long-range dependencies with $1\times1$ kernels while preserving convolutional efficiency.}
	\label{fig:coreop}
\end{figure}

As illustrated in Fig.~\ref{fig:coreop}, vision operators can be characterized along three dimensions: \textbf{window size} (unbounded, bounded, and adaptive-bounded input region per output feature), \textbf{long-range dependency} (ability to aggregate information from distant positions), and \textbf{spatial aggregation} (fixed or adaptive weighting of sampled features). Global attention~\cite{vaswani2017attention} achieves theoretically unbounded window size and true long-range modeling, but at prohibitive computational cost~\cite{pereira2024review,fan2025breaking}. Local attention~\cite{liu2021swin,liu2022swin,hu2019local,shen2021efficient,yang2021focal,chen2021crossvit} and standard convolutions are restricted to fixed local windows, while large-kernel convolutions~\cite{ding2022scaling,liu2022more,ding2024unireplknet} expand receptive fields but still rely on fixed weights within bounded windows. Deformable convolutions~\cite{dcnv1,dcnv2,dcnv3,dcnv4} improve flexibility through adaptive quadrilateral geometry defined by exactly four neighboring pixels per sampling point, yet their receptive fields remain constrained to this specific geometric form. While methods like Dynamic Region-Aware Convolution~\cite{chen2021dynamic} offer input-dependent region partitioning for filter assignment, they maintain predefined spatial partitions that cannot directly control integration region geometry. This leaves a gap for operators that provide input-dependent adjustment of both region shape and extent while maintaining computational efficiency.

To address this gap, we propose Region-Aware Deformable Convolution (RAD-Conv), which evolves from fixed-shape quadrilateral sampling to input-adaptive rectangular integration. Instead of predicting 2D offsets within fixed quadrilaterals, RAD-Conv predicts four boundary offsets (top, bottom, left, and right) per kernel element, defining axis-aligned rectangular regions whose spatial extent dynamically adapts to image content. This geometric constraint provides a principled balance: while axis-aligned rectangles cannot represent all possible shapes, they offer sufficient flexibility to model diverse spatial patterns through variable aspect ratios and extents. Unlike DCN variants that require $2K$ parameters for $K$-sized kernels (constrained to quadrilateral geometry defined by exactly four neighboring pixels), RAD-Conv enables complete control over region dimensions through just four boundary offset parameters per kernel element. Unlike Dynamic Region-Aware Convolution~\cite{chen2021dynamic}, which partitions fixed regions for filter assignment, RAD-Conv directly defines the integration region itself, allowing the receptive field to dynamically conform to object structures. Features are aggregated over these continuous regions through exact integration as proposed in~\cite{jiang2018acquisition}, enabling adaptive capture of variable spatial coverage while decoupling receptive field geometry from kernel dimensions. Though region integration increases per-layer computation compared to DCN variants, RAD-Conv can reduce the network depth required for large receptive fields, suggesting potential architectural efficiency gains in deep networks. Conceptually, RAD-Conv positions itself between deformable convolutions and attention mechanisms: it preserves the inductive biases of convolutions while approaching attention's spatial adaptability within a practical geometric constraint that balances flexibility and efficiency.

	\vspace{-0.1cm}
\section{Related Works}

\paragraph{Object Detectors:}
The trajectory of object detection has evolved from anchor-based frameworks~\cite{erhan2014scalable,szegedy2014scalable,liu2016ssd,ren2015faster,lin2017focal} to anchor-free alternatives. Methods like Faster R-CNN~\cite{ren2015faster} and RetinaNet~\cite{lin2017focal} formulate object hypotheses through fixed anchor boxes across spatial positions, scales, and aspect ratios. Each anchor location predicts offsets to adjust box coordinates via differentiable transformation, with ground-truth assignment using intersection-over-union thresholds~\cite{girshick2014rich}. However, fixed anchor configurations require careful hyperparameter tuning for optimal performance across diverse object proportions~\cite{zhang2020bridging}.

Anchor-free alternatives, such as FCOS~\cite{tian2019fcos}, CornerNet~\cite{law2018cornernet}, and CenterNet~\cite{duan2019centernet}, eliminate fixed anchors by using each feature map pixel as a reference point that regresses distances to object boundaries. This approach simplifies feature-to-boundary mapping without hand-crafted priors while maintaining high detection accuracy. RAD-Conv repurposes FCOS's boundary offset prediction mechanism in feature map space, where offsets define integration regions relative to each kernel element's position rather than absolute bounding boxes in input space. Specifically, for each kernel element at position $(i,j)$ in the feature map, RAD-Conv predicts boundary offsets that determine a rectangular region in the input feature space, whose features are then integrated to produce the output value at $(i,j)$. This enables input-adaptive receptive fields within standard convolutional networks while maintaining architectural compatibility.

Sparse methods like DETR~\cite{carion2020end,zhao2024detrs,zhu2020deformable} use global attention with learnable queries to predict object sets via bipartite matching. Dense frameworks like YOLOv8~\cite{yolov8_ultralytics} dominate real-time applications with efficient high-resolution processing, while hybrid approaches~\cite{zhang2022dino} optimize speed-accuracy trade-offs.

\paragraph{Core Vision Operators:}
The evolution of vision operators has progressed from rigid local operations toward adaptive mechanisms capable of modeling both fine-grained structure and global context. Standard convolutions~\cite{lecun1998gradient,krizhevsky2012imagenet} are constrained by small fixed kernels, while depthwise separable convolutions~\cite{howard2017mobilenets,sandler2018mobilenetv2,chollet2017xception} improve efficiency without expanding spatial coverage. Dilated convolutions~\cite{chen2014semantic,yu2015multi} expand receptive fields through strategic gaps between kernel elements.

Large-kernel convolutions in ConvNeXt~\cite{liu2022convnet} and its successors (RepLKNet~\cite{ding2022scaling}, SLaK~\cite{liu2022more}, UniRepLKNet~\cite{ding2024unireplknet}) demonstrated that kernels up to 51$\times$51 can efficiently capture broad contextual information. Deformable convolutions~\cite{dcnv1,dcnv2,dcnv3,dcnv4} introduced greater spatial adaptability by predicting input-dependent offsets that reshape receptive fields to match object geometry through continuous bilinear interpolation.

Attention mechanisms~\cite{vaswani2017attention,dosovitskiy2020image} provide flexible long-range modeling but suffer from quadratic computational complexity. Window-based attention~\cite{liu2021swin,liu2022swin,shen2021efficient,yang2021focal,chen2021crossvit} restricts computation to local windows, while deformable attention variants~\cite{xia2022vision,xia2023dat++,zhu2020deformable} improve flexibility through sparse key-point sampling.

Recent works have explored more sophisticated spatial modeling mechanisms. Chen et al.~\cite{chen2021dynamic} introduced Dynamic Region-Aware Convolution (DRConv), which automatically assigns filters to spatial regions through a learnable guided mask that partitions the feature space into regions sharing the same filter. While DRConv improves semantic representation by adapting filter assignment to spatial patterns, it maintains fixed spatial partitions that cannot dynamically adjust region geometry to input content. In contrast, RAD-Conv directly defines the spatial integration region itself through four boundary offsets per kernel element, enabling axis-aligned rectangular regions that dynamically adapt to content structure. This approach achieves practical geometric adaptability with minimal parameter overhead (4 offsets versus DRConv's region partitioning mechanism), while maintaining compatibility with standard convolutional pipelines. While DCNv3 provides sub-pixel accuracy through bilinear interpolation~\cite{dcnv3}, its receptive field geometry remains constrained to fixed-shape quadrilaterals defined by exactly four neighboring pixels per sampling point. RAD-Conv extends this capability by enabling input-dependent adjustment of region geometry (extent and aspect ratio), allowing the receptive field to dynamically conform to content structure while reducing the network depth required for large receptive fields. Though region integration increases per-layer computation compared to DCNv3, RAD-Conv reduces the network depth required for large receptive fields, suggesting potential architectural efficiency gains in deep networks.

	\vspace{-0.1cm}
\section{Proposed Method}
\begin{figure*}[t]
	\centering
	\includegraphics[width=\linewidth]{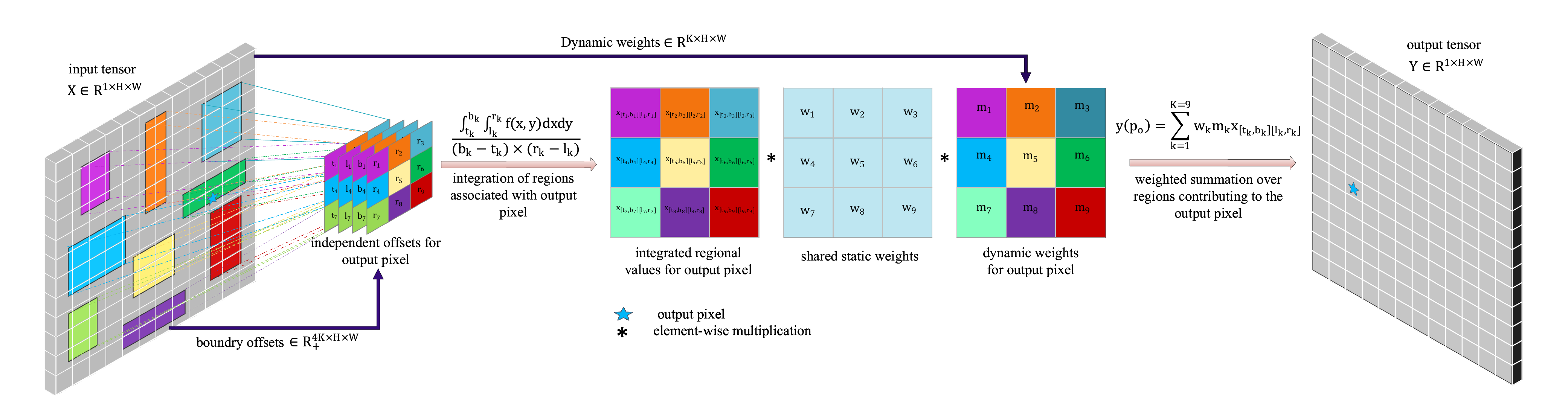}
	\caption{Region-Aware Deformable Convolution (RAD-Conv) processing flow. Given an input feature map $\mathbf{X} \in \mathbb{R}^{1\times H\times W}$ (gray grid), RAD-Conv computes a boundary offset tensor $\Delta \in \mathbb{R_+}^{4K\times H\times W}$ and a dynamic weight tensor $\mathbf{M} \in \mathbb{R}^{K\times H\times W}$, where $K=9$ corresponds to the $3\times 3$ kernel size. Unlike deformable convolutions that predict 2D offsets for $K$ sampling points within fixed quadrilaterals, RAD-Conv predicts four boundary offsets (top, bottom, left, right) per kernel element to define axis-aligned rectangular integration regions. At each spatial location, each of the $K$ kernel elements independently defines its integration region; for example, the second kernel element ($k=2$) has distinct boundary offsets and weight parameters from others. Each colored rectangular box, indexed by $k$, represents one integration region defined by its four boundary offsets. Features within each continuous region are integrated through exact spatial integration, scaled by the corresponding dynamic weight, multiplied by the shared convolutional kernel weights, and summed across all $K$ regions to produce the output feature map $\mathbf{Y} \in \mathbb{R}^{1\times H\times W}$. This geometric formulation enables complete control over region dimensions (width, height, position) through just $4K$ offset parameters, allowing receptive fields to dynamically adapt to content structure while maintaining compatibility with standard convolutional pipelines.}
	\label{fig:RAD-Conv}
\end{figure*}
\subsection{Revisiting FCOS}
Fully Convolutional One-Stage Object Detection (FCOS)~\cite{tian2019fcos} is a representative anchor-free detector that eliminates the need for predefined anchor boxes by directly utilizing the feature map's pixel locations as spatial anchors. Given an input image, a backbone CNN generates feature maps $\mathbf{F}_i \in \mathbb{R}^{H \times W \times C}$ at layer $i$, with an effective stride $s$. Each spatial location $(x, y)$ on $\mathbf{F}_i$ serves as its own anchor point, which is mapped back to image coordinates as $(x', y') = (x \cdot s, y \cdot s)$, approximately corresponding to the center of its receptive field. Unlike traditional detectors that assign multiple predefined anchor boxes to each location, FCOS treats each pixel coordinate as a single reference point for object detection. Two prediction heads are applied on top of $\mathbf{F}_i$: one for classification and one for bounding box regression, which directly predicts distances from each anchor point to object boundaries.

During inference, each location $(x, y)$ produces a classification score $p_{x,y}$ and a regression vector $t_{x,y} = (l_{\text{pred}}, t_{\text{pred}}, r_{\text{pred}}, b_{\text{pred}})$, which encodes the distances from $(x', y')$ to the four sides of the bounding box. The final bounding box is decoded as
\begin{equation} \label{eq:decode_bbox}
	\begin{aligned}
		x_{\text{tl}} &= (x - l_{\text{pred}}) \cdot s, \\
		y_{\text{tl}} &= (y - t_{\text{pred}}) \cdot s, \\
		x_{\text{br}} &= (x + r_{\text{pred}}) \cdot s, \\
		y_{\text{br}} &= (y + b_{\text{pred}}) \cdot s,
	\end{aligned}
\end{equation}
where $(x_{\text{tl}}, y_{\text{tl}})$ and $(x_{\text{br}}, y_{\text{br}})$ denote the top-left and bottom-right corners of the predicted bounding box. Note that in all equations below, the symbol `$\cdot$' denotes scalar multiplication between two real numbers. To ensure positivity, the regression branch applies an exponential transformation, guaranteeing valid bounding boxes where $x_{\text{tl}} < x_{\text{br}}$ and $y_{\text{tl}} < y_{\text{br}}$ for all predictions.

The key insight from FCOS relevant to our work is its boundary offset prediction mechanism, which defines rectangular regions through four boundary distances. RAD-Conv repurposes this geometric formulation but applies it in a fundamentally different context: while FCOS uses boundary offsets to define bounding boxes in image space for detection, RAD-Conv uses them to define continuous integration regions in feature space for adaptive feature aggregation. Specifically, for each kernel element at position $(i,j)$ in the feature map, RAD-Conv predicts boundary offsets that determine a rectangular region in the input feature space, whose features are then integrated to produce the output value at $(i,j)$. This adaptation transforms object detection geometry into a feature extraction mechanism, enabling input-adaptive receptive fields within standard convolutional networks while maintaining architectural compatibility (see Fig.~\ref{fig:RAD-Conv}).
\subsection{Evolution of Deformable Convolutions} \label{subsec:deformable_convs}
Standard convolution~\cite{lecun1998gradient,krizhevsky2012imagenet} processes input $\mathbf{x}$ through two operations: (1) uniform sampling over a fixed grid $\mathcal{R}$; (2) weighted aggregation via kernel $\mathbf{w}$. For output position $\mathbf{p}_0$, this yields:
\begin{equation} \label{standard_conv}
	\mathbf{y}(\mathbf{p}_0) = \sum_{k=1}^{K} \mathbf{w}_k \cdot \mathbf{x}(\mathbf{p}_0 + \mathbf{p}_k),
\end{equation}
where $K = |\mathcal{R}|$ denotes the total sampling points (e.g., $K=9$ for $3\times 3$ kernel with $\mathcal{R} = \{(-1,-1),\dots,(+1,+1)\}$) and $\mathbf{p}_k$ represents the $k$-th grid location. 

Deformable convolution v1 (DCNv1)~\cite{dcnv1} enhances this by introducing adaptive geometry through learnable offsets:
\begin{equation} \label{eq:dcnv1}
	\mathbf{y}(\mathbf{p}_0) = \sum_{k=1}^{K} \mathbf{w}_k \cdot \mathbf{x}(\mathbf{p}_0 + \mathbf{p}_k + \Delta \mathbf{p}_k),
\end{equation}
where $\Delta \mathbf{p}_k \in \mathbb{R}^2$ are spatial displacements predicted by a companion convolutional layer outputting $2K$ channels. Fractional offsets require bilinear interpolation:
\begin{equation} \label{eq:bilinear_interp_main}
	\mathbf{x}(\mathbf{p}) = \sum_{\mathbf{q}} G(\mathbf{q}, \mathbf{p}) \cdot \mathbf{x}(\mathbf{q})
\end{equation}
with kernel defined as:
\begin{equation} \label{eq:bilinear_interp_kernel}
	\begin{aligned}
		G(\mathbf{q}, \mathbf{p}) &= g(q_x, p_x) \cdot g(q_y, p_y), \\
		g(a, b) &= \max(0, 1 - |a - b|)
	\end{aligned}
\end{equation}
where $\mathbf{q}$ spans integer pixel coordinates. Despite this adaptability, DCNv1 remains fundamentally constrained to quadrilateral geometry defined by exactly four neighboring pixels per sampling point, as the bilinear interpolation operates over a fixed 2×2 grid.

Deformable convolution v2 (DCNv2)~\cite{dcnv2} extends DCNv1~\cite{dcnv1} by incorporating amplitude modulation:
\begin{equation} \label{eq:dcnv2}
	\mathbf{y}(\mathbf{p}_0) = \sum_{k=1}^{K} \mathbf{w}_k \cdot \mathbf{m}_k \cdot \mathbf{x}(\mathbf{p}_0 + \mathbf{p}_k + \Delta \mathbf{p}_k),
\end{equation}
where $\mathbf{m}_k = \sigma(\mathbf{m}_k^*) \in \mathbb{R}$ are sigmoid-normalized modulation scalars. Both $\Delta \mathbf{p}_k$ and raw modulation values $\mathbf{m}_k^*$ are dynamically predicted per location via a convolutional layer producing $3K$ channels ($2K$ for offsets, $K$ for modulation). While DCNv2 improves feature alignment, it maintains the same quadrilateral geometric constraint as DCNv1.

DCNv3~\cite{dcnv3} advances DCNv2~\cite{dcnv2} with two innovations: (1) replacing sigmoid with softmax normalization over sampling points; (2) partitioning channels into $G$ specialized groups. Its formulation is:
\begin{equation} \label{eq:dcnv3}
	\mathbf{y}(\mathbf{p}_0) = \sum_{g=1}^{G} \sum_{k=1}^{K} \mathbf{w}_g \cdot \mathbf{m}_{gk} \cdot \mathbf{x}_g(\mathbf{p}_0 + \mathbf{p}_k + \Delta \mathbf{p}_{gk}),
\end{equation}
where $\mathbf{w}_g \in \mathbb{R}^{(C/G) \times (C/G)}$, $\mathbf{x}_g \in \mathbb{R}^{(C/G) \times H \times W}$, $\Delta \mathbf{p}_{gk} \in \mathbb{R}^2$, and $\mathbf{m}_{gk} = \mathrm{softmax}_k(\mathbf{m}_{gk}^*)$. A dedicated convolutional layer predicts $2KG$ offset channels and $KG$ raw modulation channels. Despite these improvements, DCNv3 still requires $2K$ parameters for $K$-sized kernels to define its quadrilateral geometry, and remains constrained to the same fundamental geometric form.

DCNv4~\cite{dcnv4} overcomes DCNv3's limitations through optimized memory access patterns and removal of softmax normalization, adopting unbounded weights to enhance expressiveness. While DCNv4 achieves faster inference and better convergence, it maintains the core geometric constraint of all DCN variants: each sampling point integrates features from exactly four neighboring pixels through bilinear interpolation, limiting the ability to dynamically adjust region shape and extent according to content. This constraint becomes particularly limiting when modeling objects with irregular spatial patterns, as the quadrilateral geometry cannot independently control width and height dimensions.

The key limitation across all DCN variants is their fixed geometric form: despite adaptive positioning, each sampling point is constrained to a quadrilateral defined by exactly four neighboring pixels. This requires $2K$ parameters for $K$-sized kernels while still restricting the receptive field geometry. RAD-Conv addresses this fundamental constraint by rethinking how regions are defined, rather than predicting offsets for multiple sampling points within a fixed geometric form, RAD-Conv predicts just four boundary offsets to define axis-aligned rectangular regions (see Fig.~\ref{fig:RAD-Conv}). This simple yet powerful change enables complete control over region dimensions through only four parameters regardless of kernel size, allowing receptive fields to dynamically conform to object structures while maintaining computational efficiency.
\subsection{Region-Aware Deformable Convolutions (RAD-Conv)}
\paragraph{Forward propagation:}
Building upon the deformable convolution framework (DCNv1 \cite{dcnv1} to DCNv4 \cite{dcnv4}), which enables adaptive sampling through dynamic offsets and modulation, we identify a fundamental geometric limitation: despite operating in continuous space through bilinear interpolation, all existing DCN variants remain constrained to quadrilateral geometry defined by exactly four neighboring pixels per sampling point. As detailed in Subsection~\ref{subsec:deformable_convs}, while Eq.~\eqref{eq:bilinear_interp_main} computes values at continuous coordinates, the receptive field geometry per sampling point remains fixed to a quadrilateral form.

To overcome this geometric constraint, we fundamentally rethink how receptive fields are defined. Instead of predicting 2D offsets for multiple sampling points within a fixed geometric form, we predict four boundary offsets per kernel element to define axis-aligned rectangular regions. Inspired by FCOS's bounding box parameterization \cite{tian2019fcos} and Eq.~\eqref{eq:decode_bbox}, this approach enables complete control over region dimensions (width, height, position) through a coherent geometric parameterization. Following FCOS's approach to ensure valid region definitions, we constrain all predicted boundary offsets to be positive, which guarantees proper spatial ordering where $t_{gk} < b_{gk}$ and $l_{gk} < r_{gk}$ for all predictions. The formulation is:
\begin{equation} \label{eq:our_method}
	\mathbf{y}(\mathbf{p}_0) = \sum_{g=1}^{G} \sum_{k=1}^{K} \mathbf{w}_g \cdot \mathbf{m}_{gk} \cdot \mathbf{x}_{[t_{gk},b_{gk}][l_{gk},r_{gk}]},
\end{equation}
\begin{equation} \label{eq:expansion}
	\begin{aligned}
		t_{\text{gk}} &= (\mathbf{p}_0 + \mathbf{p}_{gk} - \Delta t^{\text{pred}}_{gk}), \\
		b_{\text{gk}} &= (\mathbf{p}_0 + \mathbf{p}_{gk} - \Delta b^{\text{pred}}_{gk}), \\
		l_{\text{gk}} &= (\mathbf{p}_0 + \mathbf{p}_{gk} + \Delta l^{\text{pred}}_{gk}), \\
		r_{\text{gk}} &= (\mathbf{p}_0 + \mathbf{p}_{gk} + \Delta r^{\text{pred}}_{gk}),
	\end{aligned}
\end{equation}
\begin{equation} \label{eq:region_integral}
	\begin{aligned}
		\mathbf{x}_{[t_{gk},b_{gk}][l_{gk},r_{gk}]} = \frac{ \displaystyle\int_{t_{gk}}^{b_{gk}}\hspace{-0.2cm}\int_{l_{gk}}^{r_{gk}}\!f(x,y)\,dxdy }{(r_{gk}-l_{gk}) \times (b_{gk}-t_{gk})}
	\end{aligned}
\end{equation}
Unlike DCN variants~\cite{dcnv1,dcnv2,dcnv3,dcnv4} that define receptive fields through multiple sampling points with quadrilateral geometry, RAD-Conv operates on axis-aligned rectangular regions $[l_{gk}, r_{gk}] \times [t_{gk}, b_{gk}]$ defined by four boundary offsets per kernel element. Here, $f(x,y)$ represents the interpolated feature value at continuous spatial coordinates $(x,y)$, and the integral computes the spatial average of all pixels lying within this rectangular box. The integral is computed analytically following the Precise RoI Pooling approach~\cite{jiang2018acquisition}, which avoids coordinate quantization by directly calculating the two-order integral based on the continuous feature map.

This geometric formulation provides a critical advantage over DCN variants: while both methods use parameter counts proportional to kernel size (DCN with $2K$ offset parameters versus RAD-Conv's $4K$ for $K$-sized kernels), RAD-Conv delivers significantly greater geometric flexibility per parameter. Specifically, it enables independent control of region width and height, which is impossible in DCN's quadrilateral geometry, allowing the receptive field to dynamically adjust its aspect ratio to match content structure while maintaining compatibility with standard convolutional pipelines. As illustrated in Figure~\ref{fig:RAD-Conv}, each of the $K$ kernel elements independently defines its integration region, with these regions spanning multiple input pixels whose contributions are integrated, scaled, and combined to produce the output.

Our approach fundamentally changes how convolutional layers access spatial information. Unlike standard CNNs and DCNs where receptive field geometry is constrained by the sampling pattern, RAD-Conv empowers each layer to dynamically define its own spatial coverage through rectangular regions with adjustable aspect ratios. This capability enables efficient processing of complex spatial patterns while maintaining architectural simplicity. RAD-Conv decouples receptive field geometry from kernel dimensions, approaching the spatial adaptability of attention mechanisms within a practical geometric constraint while retaining the efficiency of local convolutional operations.

\paragraph{Backward propagation:}
The continuous nature of our region integration enables precise gradient flow during training. The partial derivative of $\mathbf{x}_{[t_{gk},b_{gk}][l_{gk},r_{gk}]}$ \wrt $l_{gk}$ follows a derivation inspired by spatial gradient computation in object detection frameworks~\cite{jiang2018acquisition}, and is computed as:
\begin{equation}
	\begin{split}
		\frac{\partial \mathbf{x}_{[t_{gk},b_{gk}][l_{gk},r_{gk}]}}{\partial l_{gk}} = 
		\frac{\mathbf{x}_{[t_{gk},b_{gk}][l_{gk},r_{gk}]}}{r_{gk}-l_{gk}} \\
		- \frac{\int_{t_{gk}}^{b_{gk}}f(l_{gk},y)\,dy}{(r_{gk}-l_{gk}) \times (b_{gk}-t_{gk})},
	\end{split}
	\label{eqn:backpro}
\end{equation}
This derivative reveals two critical effects of boundary movement: (1) the first term $\frac{\mathbf{x}_{[t_{gk},b_{gk}][l_{gk},r_{gk}]}}{r_{gk}-l_{gk}}$ accounts for how changing the left boundary alters the region's area normalization, and (2) the second term $\frac{\int_{t_{gk}}^{b_{gk}}f(l_{gk},y)\,dy}{(r_{gk}-l_{gk}) \times (b_{gk}-t_{gk})}$ represents the feature values along the shifting left boundary that enter or exit the integration domain. When $l_{gk}$ increases (moving rightward), the gradient appropriately reduces the influence of features previously included in the region, while a decreasing $l_{gk}$ (moving leftward) increases the contribution of newly incorporated features. The partial derivatives with respect to other coordinates follow analogous principles.

Unlike DCN variants that use $2K$ offsets to shift $K$ sampling points (each constrained to bilinear interpolation over exactly four neighboring pixels), RAD-Conv uses $4K$ offsets to define $K$ rectangular regions with dynamically adjustable dimensions. This geometric formulation provides complete control over each kernel element's receptive field, enabling independent adjustment of width, height, and position, through a coherent parameterization that maintains gradient stability during optimization. The network learns to predict region boundaries that optimally balance spatial coverage and feature alignment for the given task, with the integration-based formulation ensuring smooth gradient flow throughout training.

\subsection{Window Size and Long-Range Dependency Modeling of Vision Operators}
To precisely characterize RAD-Conv's contribution, we compare vision operators in terms of window size and capability for capturing long-range dependencies. We define \textbf{window size} as the total number of input pixels that contribute to computing a single output feature, and \textbf{long-range dependency} as the ability to aggregate information from spatially distant, semantically relevant regions regardless of separation distance. As illustrated in Figure~\ref{fig:coreop}, these properties vary significantly across different vision operators.

\textbf{Standard convolutions}, defined in Eq.~\eqref{standard_conv}, have bounded window sizes strictly determined by kernel dimensions. In the extreme case of $1\times1$ kernels, illustrated in Eq.~\eqref{eq:1x1_conv}, they cannot model any spatial relationships, fundamentally limiting their capacity for long-range dependency modeling.
\begin{equation} \label{eq:1x1_conv}
	\mathbf{y}(\mathbf{p}_0) = \mathbf{w} \cdot \mathbf{x}(\mathbf{p}_0)
\end{equation}
This limitation is visually evident in subfigure (d) of Figure~\ref{fig:coreop}, where no spatial extent is modeled.

\textbf{Large-kernel convolutions} extend the receptive field by increasing kernel size, but remain constrained by fixed spatial boundaries. While they access broader regions of the input through their large kernel size, they lack the dynamism to adapt to content-specific spatial requirements due to their fixed window size, which is constrained by kernel dimensions. This fixed-window approach becomes inefficient when distant relevant features exist outside the predefined kernel region, limiting their capacity for long-range dependency modeling. This behavior is shown in subfigure (c), where coverage is broad but static.

\textbf{Deformable convolutions} improve upon this by enabling fractional coordinate access through bilinear interpolation. However, they remain fundamentally constrained by a fixed upper bound on window size. Specifically, for a kernel with $K$ sampling points, each interpolated point requires blending values from 4 neighboring pixels, resulting in an analytical window size of $4K$. This means that even with fractional offsets, the effective number of contributing pixels remains bounded. In the extreme scenario where kernel size is $1\times1$ and group size is $1$, Eq.~\eqref{eq:dcnv3} of DCNv3 simplifies to:
\begin{equation}
	\mathbf{y}(\mathbf{p}_0) = \mathbf{w} \cdot \mathbf{m} \cdot \mathbf{x}(\mathbf{p}_0 + \Delta \mathbf{p})
\end{equation}
where predicted offsets $\Delta \mathbf{p}$ enable a single deformable layer to access spatial locations beyond the fixed grid through Eqs.~\eqref{eq:bilinear_interp_main} and \eqref{eq:bilinear_interp_kernel}. While deformable convolutions can reach distant locations due to their dynamic offsets, their window size remains bounded by $4K$, where $K$ is the kernel size. Subfigures (e) and (f) show this adaptive yet bounded aggregation.

\textbf{Attention mechanisms} offer theoretically unbounded window size, as each output feature can attend to any location in the input through the attention operation:
\begin{equation} \label{eq:attention}
	\mathrm{softmax}\left(\frac{1}{\sqrt{d}}QK^\top\right)V,
\end{equation}
where $Q, K, V \in \mathbb{R}^{N \times d}$ are query, key, and value matrices, $N$ is the sequence length (number of spatial positions), and $d$ is the feature dimension. When $N$ encompasses the entire feature map (global attention~\cite{dosovitskiy2020image}), the window size becomes unbounded, enabling genuine long-range dependency modeling (Figure~\ref{fig:coreop}a). However, this global attention suffers from quadratic computational complexity $O(N^2d)$ with respect to feature map size. Local attention variants~\cite{liu2021swin} mitigate this by restricting attention to local windows of size $w \times w$ (where $w \ll \sqrt{N}$), reducing complexity to $O(Nw^2d)$ (Figure~\ref{fig:coreop}b), but at the cost of sacrificing genuine long-range dependency modeling capability.

\textbf{RAD-Conv} offers a principled balance between these extremes through continuous region integration. Even in the extreme case of $1\times1$ kernel with single-group formulation:
\begin{equation} \label{eqn:1x1}
	\mathbf{y}(\mathbf{p}_0) = \mathbf{w} \cdot \mathbf{m} \cdot \mathbf{x}_{[t,b][l,r]}
\end{equation}
the method performs integration over the continuous region $[l, r] \times [t, b]$ as defined in Eq.~\eqref{eq:region_integral}, where region boundaries are dynamically determined by input-dependent offsets. Unlike deformable convolutions with their $4K$ window size bound, RAD-Conv's window size is adaptively bounded up to image dimensions (e.g., $H \times W$ pixels), decoupling receptive field size from kernel dimensions. For a complete layer with $K$ kernel elements and $N$ output positions, RAD-Conv's computational complexity is $O(K \times N \times R^2)$ where $R$ is the side length of the integration region. This is significantly more efficient than global attention's $O(N^2d)$ when $R^2 \ll N$ (which typically holds in practice, as $R$ is constrained by feature map resolution while $N$ grows quadratically with spatial dimensions). As shown in Figure~\ref{fig:coreop}(g,h), RAD-Conv enables adaptive spatial modeling that can capture long-range dependencies while maintaining computational efficiency. A single RAD-Conv layer can achieve coverage requiring multiple stacked DCN layers, suggesting potential architectural efficiency gains in deep networks.

	\vspace{-0.1cm}
\section{Conclusion}
\vspace{-0.2cm}

We presented Region-Aware Deformable Convolution (RAD-Conv), a novel approach that extends deformable convolution by evolving from quadrilateral geometry defined by multiple sampling points to input-adaptive rectangular integration. RAD-Conv fundamentally advances spatial modeling by enabling each kernel element to dynamically adjust both the extent and aspect ratio of its receptive field according to input content, rather than being constrained by fixed geometric forms. Unlike deformable convolution variants that require $2K$ parameters for $K$-sized kernels to adjust quadrilateral geometry, RAD-Conv achieves complete control over region dimensions through just $4K$ boundary offset parameters, enabling independent adjustment of width, height, and position. This geometric formulation allows layers to capture long-range dependencies with adaptively bounded window size up to image dimensions, even with $1\times1$ kernels. RAD-Conv decouples receptive field geometry from kernel dimensions, allowing single-layer receptive fields to dynamically conform to content structure while maintaining compatibility with standard convolutional pipelines. This parameter-efficient geometric representation provides a principled balance between spatial adaptability and computational efficiency, offering a practical path toward more expressive convolutional networks. Future work will empirically validate these capabilities across vision tasks and explore applications where precise geometric modeling of spatial patterns is critical.

	{\small
		\bibliographystyle{IEEEtran} 
		\bibliography{main}
	}
	
\end{document}